\documentclass[10pt,twocolumn,letterpaper]{article}

\usepackage{cvpr}
\usepackage{times}
\usepackage{epsfig}
\usepackage{graphicx}
\usepackage{amsmath}
\usepackage{amssymb}

\usepackage{times}  
\usepackage{helvet}  
\usepackage{courier}  
\usepackage{url}  
\usepackage{graphicx}  
\usepackage{color}
\usepackage{times}
\usepackage{epsfig}
\usepackage{graphicx}
\usepackage{amsmath}
\usepackage{amssymb}
\usepackage{algorithm}
\usepackage[noend]{algpseudocode}
\usepackage{colortbl,bbm}
\usepackage{enumitem}

\usepackage{multirow}

\newcommand{\Xmat}[0]{{{\bf X}}}

\newcommand{\av}[0]{{\boldsymbol{a}}}

\newcommand{\hv}[0]{{\boldsymbol{h}}\xspace}

\newcommand{\vv}{\boldsymbol{v}}

\newcommand{\xv}{\boldsymbol{x}}

\newcommand{\zv}{\boldsymbol{z}}

\usepackage[pagebackref=true,breaklinks=true,letterpaper=true,colorlinks,bookmarks=false]{hyperref}
\cvprfinalcopy 

\def\cvprPaperID{0025} 

\begin{document}
	
	\title{Generative Model for Zero-Shot Sketch-Based Image Retrieval}
	
	\author{Vinay Kumar Verma*, Aakansha Mishra$^{\ddag}$, Ashish Mishra$^{\dag}$ and Piyush Rai*\\
		$^*$IIT-Kanpur, $^{\ddag}$IIT-Guwahati, $^{\dag}$IIT-Madras\\
		{\tt\small vkverma@cse.iitk.ac.in, ak.kkb@iitg.ac.in, mishra@iitm.ac.in,piyush@cse.iitk.ac.in}
	}
	\maketitle

	\begin{abstract}
		We present a probabilistic model for Sketch-Based Image Retrieval (SBIR) where, at retrieval time, we are given sketches from novel classes, that were not present at training time. Existing SBIR methods, most of which rely on learning class-wise correspondences between sketches and images, typically work well only for previously seen sketch classes, and result in poor retrieval performance on novel classes. To address this, we propose a generative model that learns to generate images, conditioned on a given novel class sketch. This enables us to reduce the SBIR problem to a standard image-to-image search problem. Our model is based on an inverse auto-regressive flow based variational autoencoder, with a feedback mechanism to ensure robust image generation. We evaluate our model on two very challenging datasets, Sketchy, and TU Berlin, with novel train-test split. The proposed approach significantly outperforms various baselines on both the datasets.
	\end{abstract}
	\vspace{-15pt}
	\section{Introduction}
	The commonly used approaches to search for an image from a database of images are: (1) Text-based image retrieval, in which we search for an image using a text-based query and (2) Content-based image retrieval (CBIR), in which a related image is used as a query image. Image as the query has a much richer content as compared to text-based query. CBIR gives excellent search results but requires giving a \emph{real} image as the query, which may not always be possible. Often it is more convenient to draw an outline sketch of the image and use that as a query to search for the desired image(s). The retrieval of images by giving the sketch as a query is termed as sketch-based image retrieval (SBIR) \cite{conf/iccv/SBIR1,conf/cvpr/SBIR2,conf/cvpr/SBIR15,Siamese2}. The topic has drawn considerable attention recently. However, existing SBIR systems assume that the class represented by the input sketch at query time was also present in the image-sketch pairs used to train the SBIR model, and consequently, these systems suffer when the input sketch is from a previously \emph{unseen/novel} class. 
	\begin{figure}[t]
		\centering
		\includegraphics[height=5cm,width=8.5cm]{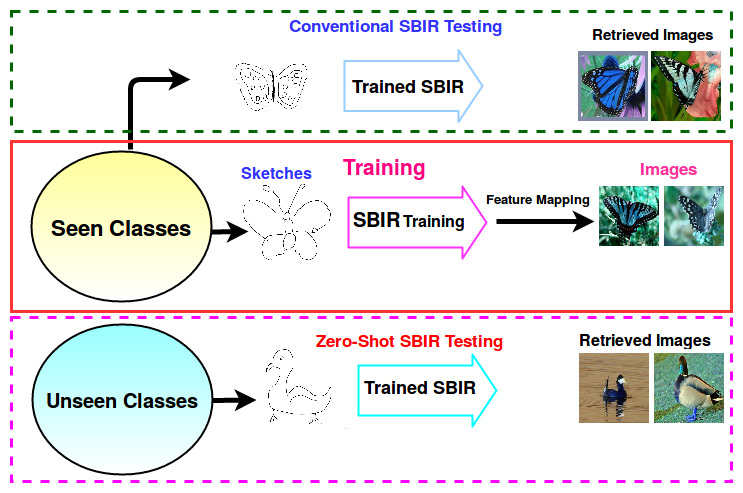}
		\caption{Illustration of Zero-Shot Sketch-Based Image Retrieval (ZS-SBIR)} 
		\label{fig:general}
		\vspace{-10pt}
	\end{figure}
	
	In this work, we present a method to handle the SBIR task for the unseen/novel class at test time. These novel classes are either absent at the training time or not used in training. This type of setup, to handle the previously unseen classes at test time is called Zero-Shot Learning (ZSL), and has been extensively investigated recently for problems, such as 	image classification \cite{vinaycvpr,ConSE,verma2017simple,akata2015evaluation}, action classification \cite{liu2011recognizing,mishrawacv}, image tagging \cite{zsl_tagging}, and visual question answering \cite{journals/corr/RamakrishnanPSM17} etc. To the best of our knowledge, the only works that have investigated SBIR in the zero-shot setting include \cite{imagehashing,ashisheccv2018}. Among these, \cite{imagehashing} used a hashing approach for the ZS-SBIR. This approach is motivated by other ZSL approaches where some side information about unseen classes is present, e.g., their textual description, word2vec or attribute based vectors are used for the knowledge transfer. Recently \cite{ashisheccv2018} proposed a vanilla conditional variational autoencoder (CVAE) architecture and adversarial autoencoder for the ZS-SBIR task.  
	
	
	
	In this paper, we address the drawbacks of existing approaches for SBIR to handle the retrieval of \emph{novel/unseen} class examples. We propose a \emph{conditional} generative model that can generate image features conditioned on the attributes (raw sketch or word2vec\cite{word2vec}) of a given class.  Like \cite{ashisheccv2018} we also have a generative model, but our approach is significantly different from their model which uses a standard conditional VAE. In contrast, our proposed approach is built upon the \emph{Inverse autoregressive flow} (IAF) based variational autoencoder \cite{kingma2016improved}, with a \emph{feedback} based mechanism \cite{controlabletext}. The IAF helps to learn the complex latent-space distribution of the images while the feedback mechanism further helps in making the generated image distribution follow the original distribution more closely. The other recently proposed ZS-SBIR approach \cite{imagehashing} requires side information in the form of description of the sketch, which may not always be available. In contrast, our proposed approach requires no side information and still performs significantly better than \cite{imagehashing}. We also use a residual decoder that helps to learn a complex model with a deeper network. Notably, since we are able to generate images from any specified class, we are able to transform the zero-shot problem into a typical supervise learning problem. The main contributions of this paper can be summarized as follows:
	\begin{itemize}
		\item We propose a sketch-conditioned image generation scheme to solve the ZS-SBIR problem, using a generative model consisting of an inverse autoregressive flow based encoder.
		\item We leverage a feedback mechanism \cite{controlabletext,vinaycvpr} to encourage the synthesized distribution to be not too far from the original distribution of the observed unlabeled images.
		\item Unlike the other recently proposed approaches for ZS-SBIR~\cite{imagehashing}, even without any side information (e.g., word2vec based attributes of the classes), our method yields significantly better results as compared to \cite{imagehashing}.
	\end{itemize}
	
	\begin{figure*}[t]
		\includegraphics[height=6.0cm,width=17.5cm]{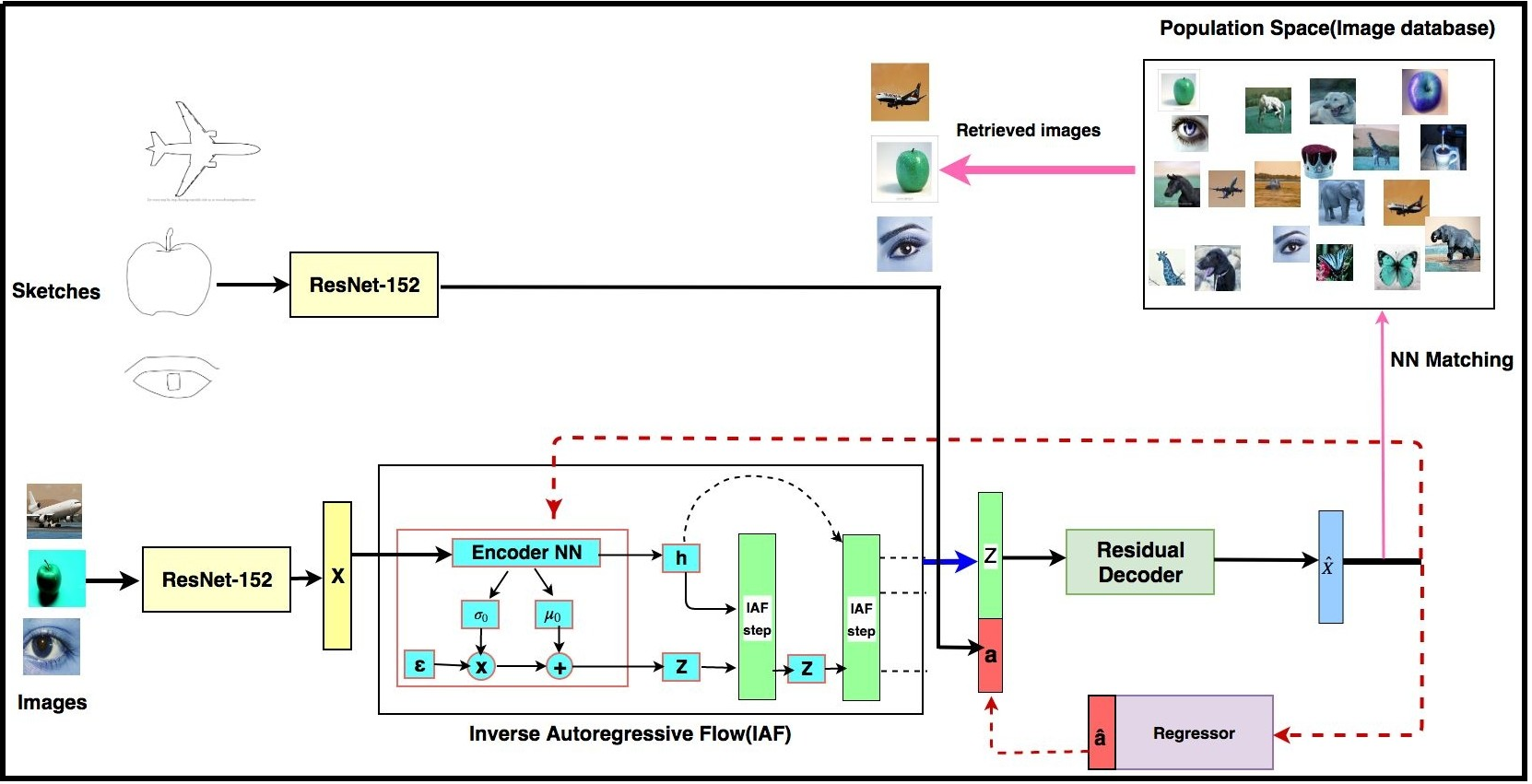}
		\caption{An illustration of our proposed model, based on the IAF architecture and feedback mechanism.} 
		\label{fig:model}
		\vspace{-10pt}
	\end{figure*}

	\section{ZS-SBIR Setting}
	In the zero-shot setting, we partition image dataset into two parts based on sketch classes. One part is the training set which has paired seen-class (S) sketches and image. The second part is test set which has unseen-class (U) sketches only (and no images). Note that the training set is essentially labelled. The training and testing set are mutually exclusive in terms of the sketch classes. In zero-shot setting, we train our model in such a way that it can generalize to unseen class sketches. The mathematical formulation of the zero-shot problem for SBIR is given below:
	
	Let $A=\{(\mathbf{x_i}^{skt},\mathbf{x_i}^{img},y_i)|y_i \in \mathcal{Y}\}$ be the triplet consisting of sketch, image and the class label, where $\mathcal{Y}$ is the set of all class labels. We partition the class labels into two disjoint set $\mathcal{Y}_{tr}$ and $\mathcal{Y}_{te}$ for train and test set respectively. Let $A_{tr}=\{\mathbf{x_i}^{skt},\mathbf{x_i}^{img},y_i|y_i \in Y_{tr}\}$ and $A_{te}=\{\mathbf{x_i}^{skt},\mathbf{x_i}^{img},y_i|y_i \in Y_{te}\}$ be the partition of $A$ into train and test set, respectively. Another assumption for the ZS-SBIR is that $A_{tr}\cap A_{te}=\emptyset$ i.e. train and test classes are disjoint. For simplicity, we will represent  $\mathbf{x}^{skt}$ as ''$\mathbf{a}$'' and $\mathbf{x}^{img}$ as ''$\mathbf{x}$'' throughout this exposition.
	
	\section{Background}
	
	As discussed earlier, our approach is based on turning the sketch-to-image search problem into an image-to-image search problem. To this end, we need a model that can generate high-quality images, given a sketch of the class representing that image. This, essentially is a conditional image generation problem. To model the complex distribution of real-world images, we leverage the inverse auto-regressive flow (IAF) based variational autoencoder~\cite{kingma2016improved}, and adapt it using a feedback mechanism to integrate the information provided by the sketch attribute. Before describing our architecture, we first provide a background of the components we build upon.
	
	\subsection{Variational Inference and Learning}
	Suppose $\xv=\{\xv^1,\cdot\cdot,\xv^N \}$ be a set of $N$ i.i.d. observations (e.g., $N$ images). Let us denote each sample by $\xv$ and assume $\zv$ be the latent variable associated with $\xv$. For a given dataset $\Xmat$, the marginal likelihood of observations is denoted as $\log p(\xv)=\sum_{i=1}^{N}\log p(\xv^i)$. The posterior over the latent variable is denoted by $q(\zv|\xv)$. We can define a variational lower bound on the marginal log-likelihood 
	\begin{equation}
	\small
	\log p(\xv)\geq E_{q(\zv|\xv)}[\log p(\xv,\zv)-\log q(\zv|\xv)] = L(\xv;\theta)
	\end{equation}
	where $p$ and $q$ are distributions whose parameters are collectively denoted by $\theta$, and $L$ is the Evidence Lower Bound (ELBO), defined as
	\begin{equation}
	L(\xv;\theta) = \log p(\xv)-D_{kl}(q(\zv|\xv)||p(\zv|\xv))
	\label{eq:elbo}
	\end{equation}
	Maximizing the lower bound $L(\xv;\theta)$ w.r.t. $\theta$ also maximizes $\log p(\xv)$ and minimizes $D_{kl}(q(\zv|\xv)||p(\zv|\xv)$, where $p(\zv|\xv)$ is the true posterior over the latent variables and $q(\zv|\xv)$ is the approximate posterior (often also called the inference network). In order to infer complex true posterior $p(\zv|\xv)$, we need to have a sufficiently expressive approximation $q(\zv|\xv)$. Normalizing Flows~\cite{germain2015made} is an idea that helps accomplish this be defining a series of transformations for a latent variable that enable learning sufficiently rich distribution for that variable.
	
	\subsection{Normalizing Flow}
	For the inference network $q(\zv|\xv)$, we need a highly flexible method that captures the complex nature of the true posterior distribution. Normalizing flow is a popular approach used for the variational inference of posterior over latent space. Normalizing flow \cite{germain2015made} depends on sequence of invertible mappings for transforming the initial probability density. Suppose $z_0$ be the initial random variable with a simple probability density function $q(\zv_0|\xv)$ and $\zv_t$ be the final output of a sequence of invertible transformations $f_t$ on $\zv_0$. $\zv_t$ can be computed as: $\zv_t=f_t(\zv_{t-1},\xv)$ $\forall t=1,\cdots,T$. If Jacobian determinant of each $f_t$ can be computed, then the  final  probability density function can be computed as:
	\[\small
	\log q(\zv_T|\xv)=\log(\zv_0|x)-\sum_{t=1}^T \log\det\left|\frac{\partial \zv_t}{\partial \zv_{t-1}}\right|
	\]
	\subsection{Inverse Autoregressive Transformations (IAF)}
	Let $\vv$ be a variable which is modeled by the Gaussian version of the autoregressive model. Suppose $[\mu(\vv),\sigma(\vv)]$ be the representation of function that maps $\vv$ to the mean $\mu$ and variance $\sigma$. Due to the autoregressive structure, the Jacobian is lower triangular matrix with zeros on the diagonal. Mean and standard deviation of $i^{th}$ element of $\vv$ are computed from $\vv_{1:i-1}$ i.e., previous elements of $\vv$.
	To sample from such a model, we use a sequence of transformations from a noise vector $\epsilon \sim N(0,I)$ to the corresponding vector $\vv$ as: $\vv_0=\mathbf{\mu}_0+\mathbf{\sigma}_0 \odot \mathbf{\epsilon}_0$ and for $i>0$ $\vv_i=\mu_i(\vv_{1:i-1})+\sigma_i(\vv_{1:i-1})\epsilon_i$. Variational inference makes sampling from posterior, such models are not interesting to be directly used for the normalizing flow. Although, the inverse transformation is interesting for normalizing flows, as long as we have $\sigma_i>0$ the transformation is one-to-one and it can be inverted as : $\mathbf{\epsilon}_i=\frac{\vv_i-\mathbf{\mu}_i(\vv_{1:i-1})}{\mathbf{\sigma}_i(y_{i:i-1})}$.
	Two key observation for IAF as follows:
	\begin{itemize}
		\item As computation of every element $\mathbf{\epsilon}_i$ does not depend on one another, inverse transformation can be parallelized $\mathbf{\epsilon}=\frac{\vv-\mathbf{\mu}(\vv)}{\mathbf{\sigma_y}}$ (subtraction and division are element-wise).
		\item Inverse autoregressive operation has a simple Jacobian determinant. It is lower triangular matrix. As an outcome, the log-determinant of Jacobian of transformation is simple to compute: $\log\det|\frac{\partial \epsilon}{\partial \vv}|=\sum_{i=1}^D -\log \mathbf{\sigma}_i(\vv)$
	\end{itemize}
	\subsection{IAF step}
	As shown in Fig.~\ref{fig:model}, the output of  initial encoder network is $\mathbf{\mu}_0$, $\mathbf{\sigma}_0$ and one extra output $\hv$ which is consider as one extra input to each subsequent step in the flow. In other word, the parameters of encoder are refined iteratively based on output of previous step $\mathbf{\mu}_0$, $\mathbf{\sigma}_0$ and $\hv$. The sampled vector from latent space of initial encoder is defined as : $\zv_0=\mathbf{\mu}_0 + \mathbf{\sigma}_0 \odot \mathbf{\epsilon}$. Where $\mathbf{\epsilon} \sim N(0,I)$. After $t$ steps the refinement of sample $\zv_0$ is recursively defined as : $\zv_t=\mathbf{\mu}_t + \mathbf{\sigma}_t \odot \zv_{t-1}$. In this sequential step the predicted posterior fits more closely to the true posterior.
	
	Finding an appropriate latent space for sampling is a crucial part of generative models as in variational autoencoder (VAE). VAE based generative models compute latent space in one step which may not be sufficient to capture a complex distribution. So the distribution of the predicted posterior and true posterior could be different with adequate margin. Whereas in IAF based variational autoencoder, predicted posterior are transformed to the true posterior using some simple sequential transformation. This sequence of simple transformation can be reduced to any complex distribution. Therefore using an auto-regressive method we can reduce the difference between the distribution of the estimated posterior and true posterior as compare to standard VAE.
	
	\section{Zero-Shot Sketch-Based Image Retrieval}
	In this section, we describe the various components of our proposed model. Again, note that the goal is to learn to generate high-quality images, given the sketch and optionally other side information (e.g., word2vec description of the class).
	\subsection{Inverse Autoregressive Flow-Based Encoder}
	Learning the complex distribution of $\zv$ in the high dimensional latent space is not feasible by a single step transformation. Therefore, in the plain VAE, the approximate posterior can be far away from the true posterior of $\zv$. IAF provides a way to learn the complex distribution by using the chain of simple transformation. The final latent variable $z$ can be given as:
	\begin{equation}
	\mathbf{z}_T=f_T(...f_2(f_1(f_0(\mathbf{z_0})))...)
	\vspace{-5pt}
	\end{equation}
	Here each $f_i$ is simple transformation function and are invertible in nature. Figure \ref{fig:model} shows the pipeline of IAF architecture.
	\subsection{VAE with feedback mechanism}
	In our model, the encoder consists of standard encoder coupled with an IAF module. The output of IAF based encoder is refinement of the latent code which is initialized by standard encoder, denoted as $p_E(z_t|x)$ with parameters $\theta_E$. The regressor output distribution is denoted as $p_R(\mathbf{a}|x)$, and the VAE loss function is given by (assuming the regressor to be fixed):
	\begin{equation}
	\small
	\begin{aligned}
	\mathcal{L}_{VAE}(\theta_E,\theta_G) &= -\mathbb{E}_{p_{E}(\mathbf{z_t}|\mathbf{x}),p(\mathbf{a}|\mathbf{x})} [\log p_{{G}}(\mathbf{x}|\mathbf{z_t},\mathbf{a})]\\ &+ \text{KL}(p_{E}(\mathbf{z_t}|\mathbf{x})||p(\mathbf{z_t}))
	\end{aligned}
	\label{eq:VAE}
	\end{equation}
	where the first term on the R.H.S. is generator's reconstruction error and the second term promotes the estimated posterior to be close to the prior.
	\subsubsection{Regressor/Cyclic-consistency Loss}
	In our proposed model, the regressor, defined by a probabilistic model $p_R(\av|\xv)$ with parameters $\theta_R$, is a feed-forward neural network that learns to project the example $\xv \in \mathbb{R}^D$ to its corresponding class-attribute vector $\av \in \mathbb{R}^L$. The objective of the regressor is to minimize the cyclic-consistency loss. The regressor is learned using two sources of data: 
	\begin{itemize} 
		\item Labeled examples $\{\xv_n,\av_n\}_{n=1}^{N_S}$ from the seen classes, on which we can define a supervised loss, given by
		\begin{equation}
		\mathcal{L}_{Sup} (\theta_R) = -\mathbb{E}_{\xv_n}[p_R(\av_n|\xv_n)]
		\end{equation}
		\item Synthesized examples $\hat{\xv}$ from the generator, for which we can define an unsupervised loss, given by 
		\begin{equation}
		\small
		\mathcal{L}_{Unsup}(\theta_R) = -\mathbb{E}_{p_{\theta_G}}(\mathbf{\hat{x}}|\mathbf{z_t})p(\mathbf{z_t})p(\mathbf{a})[p_R(\mathbf{a}|\mathbf{\hat{x}})]
		\end{equation}
		The weighted combination of supervised and unsupervised loss is defined as the overall objective to minimize the cyclic-consistency/regressor loss:
		\begin{equation}
		\vspace{-5pt}
		\min_{\theta_{R}} \mathcal{L}_R = \mathcal{L}_{Sup} + \lambda_R \cdot \mathcal{L}_{Unsup}
		\label{eq:disc_total}
		\vspace{-5pt}
		\end{equation}
	\end{itemize}
	\subsubsection{Regressor-Driven Learning}
	Regressor-Driven learning helps to minimize the cyclic-consistency loss and guide the generator to generate high-quality samples. The cyclic loss encourages the decoder/generator to generates example $\hat{\xv}$ coherent with its sketch feature vector $\av$. This is done using a loss function described below.
	
	In the first case, suppose the generator generates low-quality samples. Then the regressor will incur a high cyclic loss for these samples. In this case, the regressor assumes that it has optimal parameters and will not regress to the correct value. This loss occurs because of the bad quality samples generated by the generator. Minimizing this loss w.r.t $\theta_G$ helps generator to improve the samples quality. The objective function is given by
	\begin{equation}
	\mathcal{L}_{c}(\theta_{G}) = -\mathbb{E}_{p_G(\mathbf{\hat{x}}|\mathbf{z_t},\mathbf{a})p(\mathbf{z_t})p(\mathbf{a})}[\log p_{R}(\mathbf{a}|\hat{\xv})]
	\vspace{-5pt}
	\end{equation}
	The other loss which acts as a regularizer that encourages the generator to generate a good class-specific sample even from a random $\mathbf{z_t}$ drawn from the prior distribution $p(\mathbf{z_t})$ and combined with the sketch from $p(\av)$ is
	\begin{equation}
	\mathcal{L}_{Reg}(\theta_G) = -\mathbb{E}_{p(\mathbf{z_t})p(\mathbf{a})}[\log p_G(\mathbf{\hat{x}}|\mathbf{z_t},\mathbf{a})]
	\vspace{-5pt}
	\end{equation}
	The above two loss functions help us increase the coherence of $\mathbf{\hat{x}} \sim p_G(\mathbf{\hat{x}}|\mathbf{z},\mathbf{a})$ with class-attribute $\mathbf{a}$. A third loss function is used to ensure that the sampling distribution $p(\mathbf{z_t})$ and the distribution obtained from the generated examples $p_{E}(\mathbf{z_t}|\mathbf{\hat{x}})$ follow the same distribution.
	\begin{equation}
	\begin{aligned}
	\mathcal{L}_{E}(\theta_G) = -\mathbb{E}_{\mathbf{\hat{x}}\sim p_G(\mathbf{\hat{x}}|\mathbf{z_t},\mathbf{a})}\text{KL}[(p_{E}(\mathbf{z_t}|\mathbf{\hat{x}})||q(\mathbf{z_t}))]
	\end{aligned}
	\vspace{-5pt}
	\end{equation}
	Hence the complete learning objective for the generator and encoder is given by,
	\begin{equation}
	\begin{aligned}
	\min_{\theta_G,\theta_E}  \mathcal{L}_{VAE} + \lambda_c \cdot \mathcal{L}_{c} + \lambda_{reg} \cdot \mathcal{L}_{Reg} + \lambda_E \cdot \mathcal{L}_E \\
	\end{aligned}
	\label{eq:EG_total}
	\vspace{-5pt}
	\end{equation}
	\subsection{Residual Decoder}
	The proposed decoder is a combination of the deep and shallow network. The deep network is responsible for the better reconstruction of visual space while shallow network reduces over-fitting. This architecture is motivated by ResNet \cite{resnet} where the network has skip connections. These skip connections provide more paths to the network for information propagation. While some paths are deeper, others are shallow \cite{he2016identity}. If in the deeper path the gradient vanishing or explosion problem occurs, the shallow paths still work, and proper gradient flows in the backward direction. In the residual network, the output of a neural network layer is given by  $f_o(x)=f_{in}(\xv)+\xv$, (here $f_{in}(\xv)$, is the direct output), i.e., the output does not only depend on the current layer neural network, but it depends on input as well.
	\section{Related Work}
	Images have rich and vibrant content, while a sketch only provides rough information like shape and size. It is easy for a human to match the sketch from the image, but for machines, this is a very complex task. Since for an algorithm, it is very difficult to learn the features that are invariant to color, shape, size, pose, etc. The common pipeline for SBIR is to project the images and sketches in common subspace such that the same class images and sketches are close to each other on some metric space. Then any similarity metric can be used for the retrieval task.  Most of the traditional approaches for SBIR have used hand-crafted features such as gradient field HOG descriptor \cite{HOG}, SIFT \cite{sift} and SURF \cite{surf} etc. \cite{sarthak} proposed a dynamic programming based method for SBIR which is effective in translation, rotation, and scale (similarity). Recent advancement of deep learning provides an automatic feature extraction technique which learns the pose and color invariant feature.  Recently \cite{journals/ijcv/CNNinSKETCH,sketchy,imagehashing,ashisheccv2018} have used deep feature for SBIR task. Instead of finding the common subspace other approach projects the sketch space to image space or vice versa such that the information gap between the sketches and the real images are minimum \cite{hu2013performance,sarthak}. 
	
	Recently zero-shot learning drew more attention due to its capability of classifying a novel class object during the test phase. In the ZSL each class is associated with side information like description of the class, attribute or unsupervised word embedding (Word2vec \cite{word2vec}, Glove \cite{glove}, etc.). This side information of the class is called the semantic features/attributes. In ZSL, the core concept is to learn projection between class feature and side information, using labeled seen class data only. We can categories all proposed models for ZSL in three types based on projection.The most popular work learns the projection between visual space to semantic space and vice-versa \cite{xu2017transductive,akata2015evaluation,ConSE,vinayaaai,verma2017simple,mishracvpr}. Another popular approach projects the visual and semantic features in a shared subspace such that same class visual features and semantic attributes map closer, whereas different class visual features and semantic attributes are well-separated \cite{wang2017zero}.
	
	Recently generative models are emerging as the most popular approach for zero-shot image classification. This type of approach gaining popularity because of its ability to synthesize the unseen class sample and can reduce the ZSL problem to a supervised learning problem. These approach learns the data distribution based on the given conditions \cite{guo2017synthesizing,verma2017simple,vinaycvpr,xian2018feature}. Most of the previous methods for zero-shot learning are focused on image classification. However, a few models are used for zero-shot action classification, zero-shot image tagging and zero-shot multi-label learning as well \cite{lampert2014attribute,xu2015semantic,mishrawacv,xu2017transductive,zsl_tagging,uai_gaure}.
	
	Recently \cite{imagehashing,ashisheccv2018} have proposed a model for the ZS-SBIR. \cite{imagehashing} proposed a hashing based approach for the ZS-SBIR. The hashing architecture is based on the multi-model deep network. \cite{ashisheccv2018} proposed a generative model for the ZS-SBIR based on the CVAE architecture. The proposed approach is also a generative in nature based on IAF to get the improved variational inference \cite{kingma2016improved}. Here our encoder is based on the IAF architecture that learns the complex latent encoding of the input into the latent space. It can learn the complex distribution with the simple sequential transformation. Also, we are using the $\beta$-VAE \cite{betavae} architecture for the disentangled representation. The residual decoder is used that gives the better generation of the sample because it can flow the gradient with the deeper layers. In the proposed approach the external feedback mechanism provides the feedback to the encoder about the generation quality. Hence the generator has better guidance for generating the robust sample.
	\section{Experiments and Results}
	To show the effectiveness of our proposed model we ause two challenging datasets: Sketchy \cite{sketchy} and TU-Berlin \cite{berlin}. Originally, Sketchy dataset \cite{sketchy} contains 75471 hand-drawn sketches and 12500 corresponding images from 125 classes. \cite{liu} have provided 60502 more real images from all 125 classes, which extends the original dataset. TU-Berlin extended \cite{berlin} is a large scale dataset having 20000 sketches and 204489 images from 250 different categories provided by \cite{liu,sketchnet}.
	
	The visual features for images and sketches are extracted using ResNet-152 \cite{resnet}, pretrained on ImageNet \cite{imagenet2015} dataset. The sketches and image features are extracted from the last fully connected layer. It gives 2048-dimensional feature vectors. We believe that further finetuning on this dataset on ResNet-152 architecture will give better performance. The visual features of the sketches are used as a class attributes in our proposed generative model.
	
	\subsection{Sketchy Dataset (Extended)}\label{sketch}
	For fair comparison with the recent work \cite{imagehashing,ashisheccv2018}, we have two splits of the dataset. 
	\cite{imagehashing} randomly selected 25 classes of sketches as the test set ($A_{te}$) and the remaining labeled 100 classes are used as the training set ($A_{tr}$). Here $A_{tr}\cap A_{te}=\phi$, i.e. train and test class are disjoint. We have another split of the dataset similar to \cite{ashisheccv2018}, this contains 104 classes in training, and used 21 classes images and sketches as a test set.
	
	Random split proposed by \cite{imagehashing} is not the realistic Zero-Shot setting \cite{ZSL-GBU}. Since in random split test set may have some classes that are present in the ImageNet class. Since we are using the ImageNet pre-trained model for the feature extraction and this training is done in a supervised manner. This violates the assumption that $A_{te}$ are the unseen classes. Therefore it is not the exact Zero-Shot setting. The split proposed by \cite{ashisheccv2018} is the realistic setup for ZS-SBIR, where the split is done in such a way that any of the $A_{te}$ classes are not present in the ImageNet dataset. In our setup for training, we need a paired image and sketch set. To make the paired set, we selected a random image and sketch from the same training class and paired them. This process repeated 1000 time, i.e., each class in the training set has 1000 pair of data point. We are comparing our model with the previous approach in their original setup; therefore, we are using both the split.
	\subsection{TU Berlin Dataset (Extended)}
	Similar to \cite{imagehashing} for the fair comparison randomly 30 classes are 
	selected for the $A_{te}$ and remaining 220 classes are used for the $A_{tr}$. This dataset is highly biased; few classes have large examples while few have only limited samples. In the Zero-shot setup, learning with biased data is a very hard problem. Therefore form training we removed the biases. For doing so, we are sampling the equal number of image and sketch sample pairs from each class. In the testing, we selected the class that has more than 400 samples. For making image and sketch pair, we follow the same pattern mentioned in the previous section. Here again, each class has a 1500 pair of image and sketch.
	
	\begin{table*}
		\begin{center}
			\scalebox{.8}{
				\addtolength{\tabcolsep}{16pt}
				\begin{tabular}{|l|l|c|c|c|c|}
					\hline 
					\textbf{Type} &  \multirow{2}{*}{\textbf{Method}} & \multicolumn{2}{c|}{\textbf{Sketchy Dataset}} & \multicolumn{2}{c|}{\textbf{TU Berlin Dataset}}\\
					\cline{3-6} 
					& & \textbf{Precision@100} & \textbf{mAP@all} & \textbf{Precision@100} & \textbf{mAP@all}\\
					\hline 
					\hline
					& Softmax Baseline & 0.176 & 0.099 & 0.139 & 0.083\\
					& Siamese CNN \cite{Siamese2}  & 0.183& 0.143 & 0.153 &0.122  \\
					& SaN \cite{journals/ijcv/CNNinSKETCH} & 0.129 & 0.104 & 0.112 &0.096 \\
					SBIR & GN Triplet \cite{sketchy} & 0.310 & 0.211 & 0.241 & 0.189\\
					& 3D Shape \cite{journals/corr/SBIR13} & 0.070 & 0.062 & 0.063 &0.057 \\
					& DSH (64 bits) \cite{journals/corr/LiuSSLS17} & 0.227 & 0.164 & 0.198 &0.122 \\
					\hline
					& CMT \cite{conf/nips/CMT} & 0.096 & 0.084 & 0.082 &0.065 \\
					& DeViSE \cite{conf/nips/DeviSE} & 0.078 & 0.071 & 0.075 & 0.067\\
					& SSE \cite{zhang2015zero} & 0.154 & 0.108 & 0.133 & 0.096\\
					Zero-Shot & JLSE \cite{zhang2016zero} & 0.178 & 0.126 & 0.165 &0.107\\
					& SAE \cite{zhang2015zero} & 0.302 & 0.210 & 0.210 &0.161 \\
					& DSH \cite{journals/corr/LiuSSLS17} & 0.217 & 0.165 & 0.174 & 0.139\\
					& ZSIH \cite{imagehashing} &   0.340    &  0.254       &   0.291   &  0.220    \\
					\hline
					Feedback-Auto & GZS-SBIR(Our)  & \bf{0.305}& \bf{0.253} &\bf{0.281} & \textbf{0.187}\\
					Feedback-VAE & GZS-SBIR(Our) &\bf{0.358} & \textbf{0.289} & \bf{0.334}& \bf{0.238}\\
					\hline 
				\end{tabular} 
			}
			\vspace{5pt}
			\caption{Precision@100 and mAP@all results on the traditional SBIR and ZSL method on the ZS-SBIR setup.  Feedback-Auto is the IAF autoencoder with the feedback mechanism and Feedback-VAE is the IAF-VAE with the feedback mechanism.
				\label{tab:oldsplit}}
		\end{center}
	\end{table*}
	
	\subsection{Implementation details}
	In our model, we have three components, Encoder (E), Generator (G) and Regressor (R). The encoder is based on the IAF architecture, refer to figure-\ref{fig:model}. The encoder contains the two fully connected layers of size 4096 followed by one layer that gives the $\mu$ and $\sigma$ this passed to 3 layers IAF architecture. The generator has five layers of the fully connected neural network (NN) with the residual connection. It is a combination of the deep and shallow network. Here sigmoid activation is used. All layers are of the same size of 6144. Regressor takes the reconstructed samples $\hat{x}$ and regresses the sketch. It uses the two-layer fully connected NN of size 4096. The learning rate ($\eta$) is set as a stepwise decreasing rate. Initially for the 5 epoch $\eta=0.001$ then after each 10 epoch it changed to [0.0005,0.0001,0.00001]. Here instead of $\mathcal{N}(0,I)$ prior, we found from the validation data $p\sim\mathcal{N}(0,0.005)$ gives the better performance. Also, for the ablation, we experiment with the plain autoencoder. The autoencoder used contains the same architecture as the IAF encoder with feedback connection; only the difference is that the dimension of $z$ is zero. Therefore the generated sample is deterministic and depends only on the given sketch feature $\av$.
	
	\subsection{Training and Testing}
	There are two modules in the model, IAF-VAE and regressor. We are alternately optimizing the IAF-VAE and regressor. These two module helps each other to learn the robust generator. In the VAE training, we are minimizing the loss w.r.t. $E$ and $G$'s parameters, and for regressor training, we are minimizing the regressor's loss w.r.t $R$'s parameter only. The alternate optimization is done Until convergence. The complete setup is for the zero-shot learning; therefore the testing is performed from the unseen class sketch to unseen class image. In the testing phase each $x^{skt}$ is concatenated with the $z\sim\mathcal{N}(0,0.005)$ and generate the $c$ samples using generator $G$. Now from these $c$, samples find the image that gives maximum cosine similarity. The similarity of the sample $x_i$ from the query sketch $x^{skt}$ can be given as:
	\begin{equation}\label{eq:test}
	\small
	s(\xv^{skt},\xv_i)=\max_{t=1:c} cosine\left(G(\theta(\xv^{skt}_t)),\theta(\xv_i)\right)
	\end{equation}
	Here, $\xv_i$  is the image in the query database. Equation-\ref{eq:test} is repeated for each image and find the $K$ samples with maximum similarity scores for the $top@K$ retrieval. $\theta$  is the ResNet-152 model and gives the feature vector for each image, and $G$ is the generator.
	
	\begin{figure}[!t]
		\includegraphics[scale=0.45]{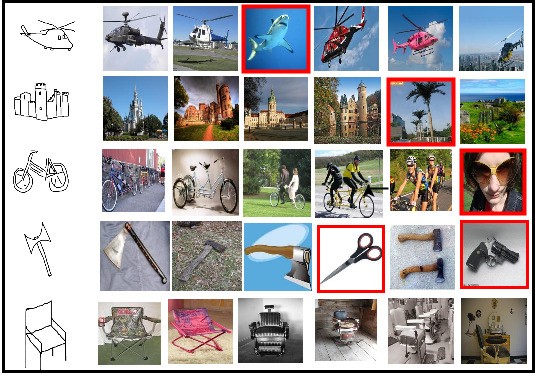}
		\caption{Top-6 retrieved images for randomly chosen five sketches for ZS-SBIR.} 
		\label{fig:retrieved}
	\end{figure}
	\subsection*{Result Analysis with existing methods}
	Since best of our knowledge, only two very recent works ZSIH \cite{imagehashing} and CVAE \cite{ashisheccv2018} have been proposed for ZS-SBIR. Therefore for supporting the performance of the proposed model, we compare the performance of our model with several other state-of-the-art. We have analyzed two types of baselines methods, 1- Sketch-Based Image Retrieval (SBIR) Baselines and 2- Zero-Shot Learning (ZSL) baselines.
	
	\subsubsection*{Sketch base image retrieval baselines (SBIR)} Several approaches have been proposed for SBIR. We compare our model with Siamese-1 \cite{conf/cvpr/Siamese1}, Siamese-2 \cite{Siamese2}, Coarse-grained triplet \cite{cgt}, Fine-grained triplet \cite{sketchy}, DSH \cite{journals/corr/LiuSSLS17}, SaN \cite{journals/ijcv/CNNinSKETCH}, GN Triplet \cite{sketchy}, 3D Shape \cite{journals/corr/SBIR13}, Siamese CNN \cite{conf/icip/SBIR16}. Since these baselines are not originally proposed for zero-shot setting, \cite{imagehashing} provides these baseline for the zero-shot setting. We have taken these baseline result directly from the paper \cite{imagehashing}. 
	\vspace{-5pt}
	\subsubsection*{Zero-Shot baselines (ZSL)} The most of the existing approaches for zero-shot learning are proposed for the zero-shot image classification. We select a set of zero-shot learning approaches as baseline to compare with our proposed model. These ZSL baseline approaches are CMT \cite{conf/nips/CMT}, DeViSE \cite{conf/nips/DeviSE}, SAE \cite{SAE2017}, SSE \cite{zhang2015zero}, ESZSL \cite{romera2015embarrassingly}, CAAE \cite{AAE}, JLSE \cite{zhang2016zero}, DAP \cite{lampert2014attribute}. The baseline results are borrowed from the \cite{ashisheccv2018,imagehashing}. We again reproduce the the baseline results reported in the table-\ref{tab:newsplit}.
	
	The recent work on the ZS-SBIR is ZSIH \cite{imagehashing} and CVAE \cite{ashisheccv2018}. ZSIH \cite{imagehashing} shows the experiment on the TU-Berlin and Sketchy dataset. They reported the result of precision@100 and mAP@all for all datasets. \cite{imagehashing} using the word2vec\cite{word2vec} as side information in their model. As mentioned earlier we are not using any side information but have significantly better result compare to ZSIH. The comparison result with the baseline and ZSIH are shown in the table-\ref{tab:oldsplit}. We can see without any side information our approach performs significantly better than all the previous approach that used the side information. Also, we experimented the proposed approach with autoencoder only and found that the proposed VAE model significantly outperforms the autoencoder model. Please refer to table-\ref{tab:oldsplit} for the more details.  
	\begin{table}[!t]
		\begin{center}
			\scalebox{.75}{
				\addtolength{\tabcolsep}{-2pt}
				\begin{tabular}{|l|l|c|c|}
					\hline
					\multicolumn{4}{|c|}{\textbf{Sketchy Dataset}} \\
					\hline 
					\textbf{Type}&{\textbf{Method}} & \multicolumn{1}{c|}{\textbf{Precision@200}} & \multicolumn{1}{c|}{\textbf{mAP@200}}\\
					\cline{2-3} 
					\hline 
					\hline
					&Baseline &  0.106 & 0.054\\
					& Siamese-1 \cite{Siamese1} &  0.243 & 0.134\\
					& Siamese-2 \cite{Siamese2} & 0.251 &  0.149 \\
					SBIR & Coarse-grained triplet \cite{cgt}& 0.169  & 0.083\\
					& Fine-grained triplet \cite{sketchy}& 0.155  & 0.081\\
					& DSH$^1$ \cite{journals/corr/LiuSSLS17} &  0.153 & 0.059 \\
					\hline
					& DAP \cite{lampert2014attribute} &0.066 & 0.022 \\
					& ESZSL \cite{romera2015embarrassingly} &0.187 & 0.117 \\
					ZS-SBIR & SAE  \cite{SAE2017} & 0.238 & 0.136 \\
					& CAAE CAAE \cite{AAE} &0.260 & 0.156 \\
					& CVAE \cite{ashisheccv2018} &0.333 & 0.225 \\
					\hline
					{Feedback-Auto} & GZS-SBIR(our) &\bf{0.288} &  \bf{0.191}\\
					{Feedback-VAE} & GZS-SBIR(our) &\bf{0.343} &  \bf{0.238}\\
					\hline 
				\end{tabular}  }
				\vspace{5pt}
				\caption{Precision@200 and mAP@200 results on the traditional SBIR and ZSL method on the ZS-SBIR setup. This table follow the realistic train-test split.}
				\label{tab:newsplit}
				\vspace{-15pt}
			\end{center}
		\end{table}
		
		Another approach CVAE \cite{ashisheccv2018} has proposed a generative model for ZS-SBIR; they showed the result on the Sketchy dataset. CVAE suggested the realistic train-test split similar to \cite{ZSL-GBU} for the ZS-SBIR. \cite{ashisheccv2018} evaluated the performance of the model over precision@200 and mAP@200 metric. We are following the same setup to compare our result with CVAE. Our result on the Sketchy-dataset shows that the proposed approach is significantly better compare to CVAE. CVAE not using any side information and without using any side information (e.g., word2vec), our method shows the $3.0\%$ and $5.8\%$  relative improvement over precision@200 and mAP@200 metric.
		
\begin{figure}[!htb]
			\centering
			\includegraphics[scale=0.16]{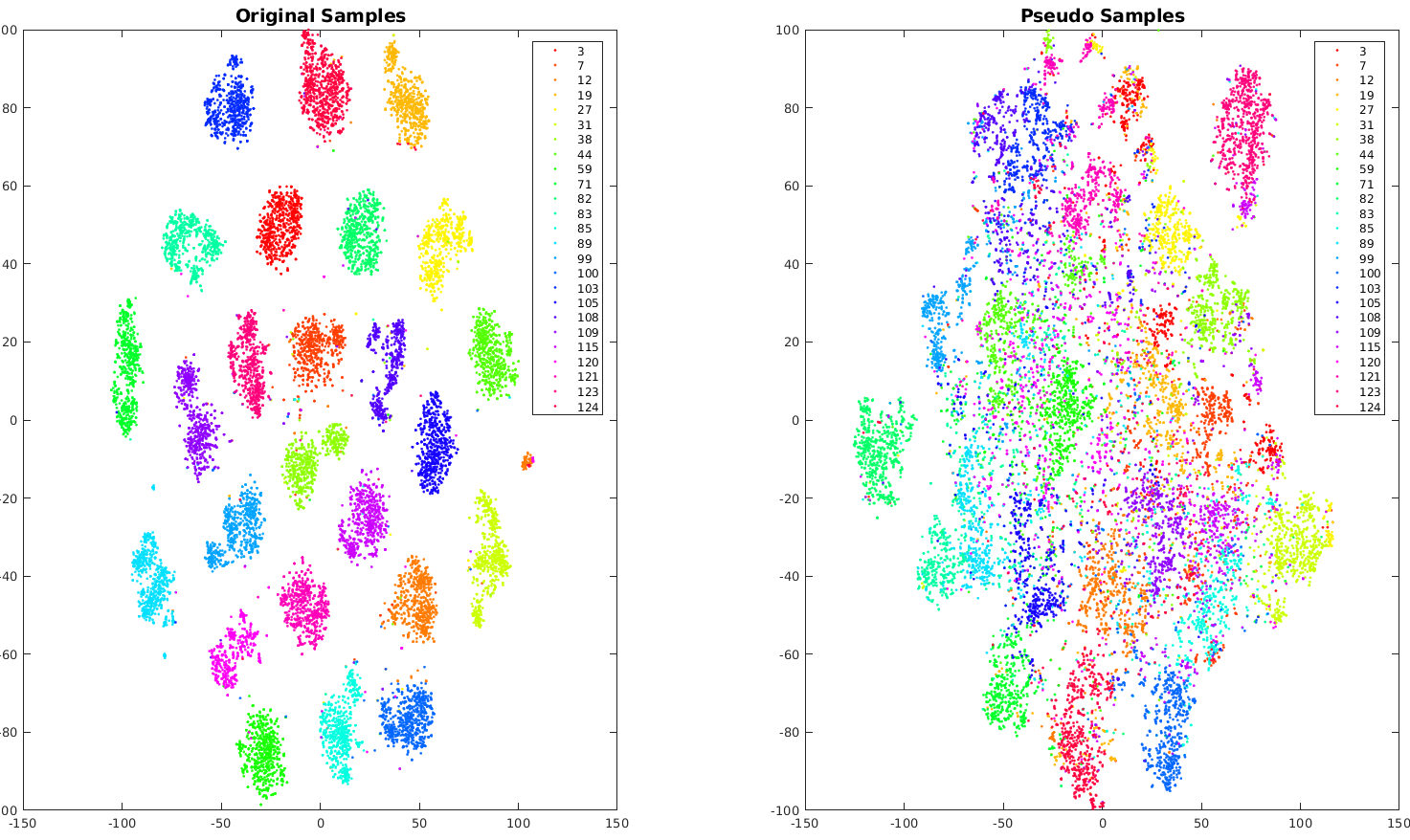}
			\caption{t-SNE plot for the original and the reconstructed sample for the sketchy dataset without using any side information.}
			\label{fig:tsne}
		\end{figure}

		In the Figure-\ref{fig:retrieved} we have illustrated the top-6 retrieved result using the unseen class sketches from the image database. Retrieved images are closely matched to the outline of the sketches. Since our model learns the mapping between sketches and images based on components. So it may retrieve some other class images which are significantly similarity with the sketch. In Figure-\ref{fig:retrieved} we can see that for helicopter sketch our model retrieve the fish because the outline of sketches of the helicopter and fish are very similar. Also, we have shown the t-SNE \cite{tsne} plot of the original data and the reconstructed data for the Sketchy \cite{sketchy} dataset. In the t-SNE plot, we can observe that the generated samples for the novel classes are not as good as the original one. But the generated sample nearly follow the same distribution as the original one. The few class samples are as good as the original samples. Please refer to figure-\ref{fig:tsne} for the t-SNE plot.
		
		\begin{figure}[b]
			\centering
			\includegraphics[scale=0.35]{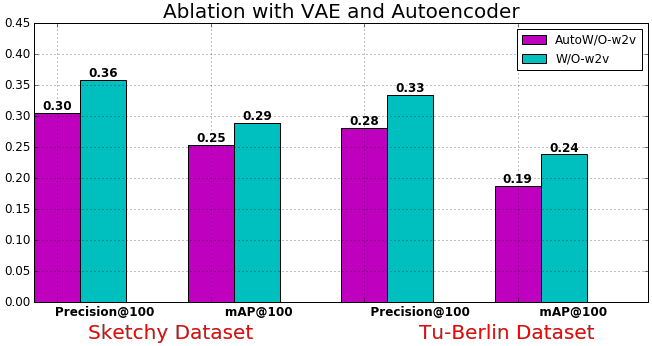}
			\caption{Ablation of our proposed VAE with the plain autoencoder without any side information. \textit{\textbf{AutoW/O-w2v}}: Autoencoder without word2vec,  \textit{\textbf{W/O-w2v}}: Proposed approach without word2vec.}
			\label{fig:ablation_autoencoder}
		\end{figure}
		\section*{Ablation Study}
		We now show the significance of the different components of the model as compared to the basic VAE model. We have found that the proposed approach outperforms by a significant margin across all the dataset. Even though without using any side information we are performing better than the previous approach \cite{imagehashing} that used the side information. In the section-[\ref{withvae}] we are showing the ablation with and without VAE. Also in section-[\ref{withiaf}] we are showing the significance of the IAF component. 
		\subsection{With/Without VAE}
		\label{withvae}
		We also perform the ablation analysis over the different component of the proposed approach. In the first experiment, we compare the performance of autoencoder with the proposed VAE architecture. We found that the proposed model with the VAE component always outperforms compare to autoencoder architecture. Using the feedback-VAE architecture on the Sketchy-dataset the model shows the $20\%$ and $16\%$ relative improvement on the precision@100 and mAP@all metric compare to plain autoencoder architecture. The similar pattern we observe for the Tu-Berlin dataset also. The feedback-VAE shows the $17.8\%$ and $26.3\%$ relative improvement over the plain autoencoder architecture on the precision@100 and mAP@all metric. Please refer to figure-\ref{fig:ablation_autoencoder} for comparison details.
		
		\subsection{With/Without IAF}
		\label{withiaf}
		We present the ablation study with IAF and without IAF component, without using any side information. We have found that if we remove the IAF component, the performance drop is significant as compared to with-IAF. For the Sketchy dataset, we reported in Table-\ref{tab:ablation} that with-IAF component, the precision@100 and mAP@all are 0.358 and 0.289, respectively. If we remove the IAF component, the performance drop is significant, and precision@100 and mAP@all are 0.313 and 0.261, respectively. Therefore we have 12.6\% and 9.7\% relative drop in the performance without-IAF. We also observed a similar pattern on the TU-Berlin dataset. Here in Table-\ref{tab:ablation} with-IAF we have 0.334 and 0.238, precision@100 and mAP@all, respectively. But if we drop the IAF component, our precision@100 and mAP@all are 0.294 and 0.198, respectively. Therefore we have 12.0\% and 16.8\% performance drop, respectively. Please refer to Table-\ref{tab:ablation} for the more details.
		
		\begin{table}[!htb]
			\begin{center}
				\scalebox{.67}{
					\begin{tabular}{|l|c|c|c|c|}
						\hline 
						\textbf{Type}  & \multicolumn{2}{c|}{\textbf{Sketchy Dataset}} & \multicolumn{2}{c|}{\textbf{TU Berlin Dataset}}\tabularnewline
						\cline{2-5} 
						& \textbf{Precision@100} & \textbf{mAP@all} & \textbf{Precision@100} & \textbf{mAP@all}\tabularnewline
						\hline 
						W/O-w2v &  0.313& 0.261 & 0.294 & 0.198 \\
						W/O-w2v+IAF &  0.358 & 0.289 & 0.334 & 0.238\\ \hline
						\textbf{Improvement (\%)} &  \textbf{12.6\%} & \textbf{9.7\%} & \textbf{12.0\%} & \textbf{16.8\%}\\
						\hline 
					\end{tabular} 
				}
				\vspace{5pt}
				\caption{\textbf{Ablation study:} Precision@100 and mAP@all results on the Sketchy and TU-Berlin dataset without-IAF and with-IAF.}
				\label{tab:ablation}
				\vspace{-5pt}
			\end{center}
		\end{table}
		
		\section{Conclusion}
		In this paper, we addressed the Zero-Shot Sketch-Based Image Retrieval problem, which is a challenging and more realistic setting as compared to the conventional SBIR. The proposed generative approach can solve the SBIR problem when the classes are growing with time, and does not require all classes to be present at the training time. We have found that the proposed approach, based on the IAF architecture with the feedback mechanism, generates high-quality samples of the novel classes. Moreover, without using any side information, our proposed generative model can retrieve novel class examples and gives state-of-art results on benchmark datasets. In this work, we assume that the test query comes from the unseen classes only. In future, it will be interesting to explore the \emph{Generalized} ZS-SBIR problem where test query can come from the seen as well as unseen classes. Also, the domain shift is a critical problem in the ZSL. It will be an interesting direction of future work to handle the domain shift for zero-shot SBIR. The recent model shows a significant improvement inj ZS-SBIR using the side information. In future, it will also be exciting to explore the model with the help of side information.
		
		\newpage
		\small
		\bibliographystyle{ieee}
		\bibliography{egbib.bib}
		
	\end{document}